\documentclass[10pt,conference,a4paper]{IEEEtran}
\IEEEoverridecommandlockouts
\usepackage{cite}
\usepackage{amsmath,amssymb,amsfonts}
\usepackage{algorithmic}
\usepackage{graphicx}
\usepackage{textcomp}
\usepackage{xcolor}
\usepackage{array,balance}
\def\BibTeX{{\rm B\kern-.05em{\sc i\kern-.025em b}\kern-.08em
    T\kern-.1667em\lower.7ex\hbox{E}\kern-.125emX}}
\begin{document}

\title{Multi-Resolution Fusion and Multi-scale Input Priors Based Crowd Counting\\
}

\author{\IEEEauthorblockN{Usman Sajid$^\dag$, Wenchi Ma$^\dag$, Guanghui Wang$^{\dag\ddag}$}
\IEEEauthorblockA{$^\dag$\textit{Department of Electrical Engineering and Computer Science,} 
\textit{University of Kansas,
Lawrence, KS, USA, 66045} \\
$^\ddag$ \textit{Department of Computer Science},
\textit{Ryerson University,
Toronto, ON, Canada M5B 2K3}\\
\{usajid, wenchima, ghwang\}@ku.edu}
}

\maketitle

\begin{abstract}
Crowd counting in still images is a challenging problem in practice due to huge crowd-density variations, large perspective changes, severe occlusion, and variable lighting conditions. The state-of-the-art patch rescaling module (PRM) based approaches prove to be very effective in improving the crowd counting performance. However, the PRM module requires an additional and compromising crowd-density classification process. To address these issues and challenges, the paper proposes a new multi-resolution fusion based end-to-end crowd counting network. It employs three deep-layers based columns/branches, each catering the respective crowd-density scale. These columns regularly fuse (share) the information with each other. The network is divided into three phases with each phase containing one or more columns. Three input priors are introduced to serve as an efficient and effective alternative to the PRM module, without requiring any additional  classification operations. Along with the final crowd count regression head, the network also contains three auxiliary crowd estimation regression heads, which are strategically placed at each phase end to boost the overall performance. Comprehensive experiments on three benchmark datasets demonstrate that the proposed approach outperforms all the state-of-the-art models under the RMSE evaluation metric. The proposed approach also has better generalization capability with the best results during the cross-dataset experiments.
\end{abstract}


\section{Introduction}
\label{intro}
Crowd counting finds a very important and integral place in the crowd analysis paradigm. Crowd gatherings are ubiquitous and bound to happen frequently at sports, musical, political, and other social events. Automated crowd counting plays an important role in handling and analyzing such events. Crowd counting is an active research area in the computer vision field due to the fact that many key challenges remain yet to be reasonably addressed, such as severe occlusion, huge crowd diversity within and across different regions in the images, and large perspective changes. Moreover, manual human based crowd counting process is  unreliable and ineffective due to the tedious and time-consuming nature of this task.

In recent years, computer vision has witnessed great developments in several sub-areas, such as image classification \cite{wu2019unsupervised}, object detection \cite{ma2020mdfn}, image translation \cite{xu2019adversarially} and face recognition \cite{cen2019dictionary}, with the introduction of convolution neural networks (CNNs). Inevitably, recent state-of-the-art crowd counting methods are overwhelmingly dominated by the CNN based approaches, which generally belong to either direct-regression (DR) \cite{wang2015deep,fu2015fast,sajid2020zoomcount} based or density-map estimation (DME) \cite{zhang2016single,cascadedmtl,sam2017switching,ranjan2018iterative,liu2018decidenet,wan2019residual,xu2019learn} based architectures. DR based methods directly regress or estimate the crowd number from the input image or patch. These methods alone do not prove effective for crowd counting due to huge crowd diversity and multi-scale variation in and across different images. The DME based methods perform crowd counting by estimating the crowd-density value per pixel. This type of approaches, in general, also tend to struggle against the above stated major issues and challenges.

Multi-column or multi-regressor CNN based architectures \cite{zhang2016single,sam2017switching,ranjan2018iterative,cascadedmtl} have proved to be very effective for crowd counting task. MCNN \cite{zhang2016single} is a state-of-the-art three-column density-map estimation based end-to-end crowd counting network, where each CNN based column specializes in handling the specific crowd-density level. At the end of this network, all columns are merged together to yield the crowd estimate after remaining processing. Similarly, multi-column based architectures \cite{sam2017switching,liu2018decidenet} utilize multiple specialized crowd count regressors to cope with multiple crowd-density scales separately. For example, Switch-CNN \cite{sam2017switching}, a density-map estimation based network, consists of a CNN based switch classifier that routes the input image or patch to one of three crowd count regressors, where each regressor deals with specific crowd level. In addition, many single-column or single-regressor based architectures \cite{li2018csrnet, shami2018people} have also been proposed to address the crowd counting issues and challenges. These methods produce promising results, but still lack the generalization ability for crowd estimation, ranging from low to high crowd-density. 

Recently, Sajid et al. \cite{sajid2020zoomcount,sajid2020plug} observed that suitable rescaling (down-, no-, or up-scaling) of the input image or patch, according to its crowd density level (low-, medium-, or high-crowd), gives more effective results as compared to the multi-column or multi-regressor based methods. Based on this observation, they also designed a patch rescaling module (PRM) \cite{sajid2020plug} that rescales the input image or patch accordingly based on its crowd-density class label. Although the PRM based single-column proposed schemes \cite{sajid2020plug} empirically prove their observation to be imperative and effective, the PRM module does not fully capitalize on it and thus limits the efficacy of this observation. First, it requires the crowd-density classification label of the original input patch. This additional classification process comes up with its own inaccuracies \cite{sajid2020zoomcount,sajid2020plug} that compromises the subsequent crowd counting process. Second, the PRM module selects only one of three available recaling operations (down-, no-, or up-scaling) for any given input patch. This limits the overall effectiveness and improvement of the PRM module and only utilize the deployed observation partially. Contrary to only using the single rescaling for the input patch, we empirically observed that using all three rescaled versions of the input patch with feature-level fusion or sharing gives much better performance. Consequently, it also eliminates the need for any crowd-density classification process for the original input patch. To this end, we aim to achieve the following two objectives in this work:

\begin{itemize}\setlength\itemsep{-0.3em}
  \item Better generalization ability: Design a multi-column crowd counting method with better generalization ability towards huge crowd variations.
\\  
  \item Effective input priors: Utilize the input patch rescaling based effective observation \cite{sajid2020zoomcount,sajid2020plug} (as discussed above) without performing any expensive and compromising crowd-density classification process, and also use all three crowd-density levels (low-, medium, and high-crowd) in a more effective manner than the PRM module \cite{sajid2020plug}.
\\  
\end{itemize}

\begin{figure*}

	\begin{minipage}[b]{1.0\textwidth}
		\begin{center}
			\centerline{\includegraphics[width=1.0\textwidth]{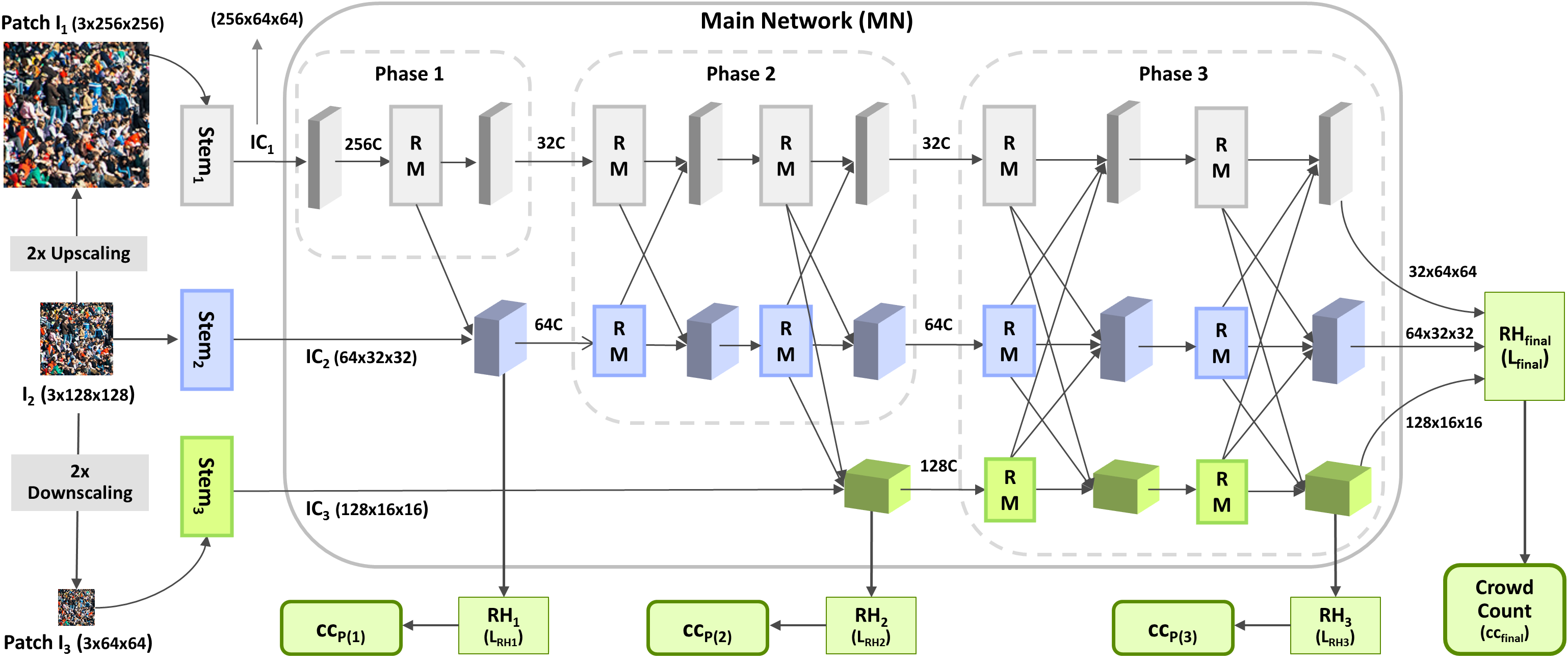}}
		\end{center}
	\end{minipage}		
	\caption{The proposed network. The original $128\times128$ input patch ($I_2$ or $P$) is used to produce the new up-scaled ($I_1$) and down-scaled ($I_3$) input priors, which go through their respective stems ($stem_1,stem_2,stem_3$). The resultant initial channels ($IC_1,IC_2,IC_3$) then pass through the phase-based main network, containing three deep columns/branches with the residual modules ($RM$). Multi-resolution fusion regularly occurs between these columns, followed by passing through the auxiliary ($RH_1,RH_2,RH_3$) and the final ($RH_{final}$) crowd regression heads to yield the respective crowd counts ($cc_{p(1)},cc_{p(2)},cc_{p(3)},$ and $cc_{final}$). The final crowd count for the input patch ($I_2$) is the weighted average of these crowd estimates. The MN maintains the channels ($C$) resolution throughout each column. (In this paper, we used both terms ($I_2$ and $P$) interchangeably for the same original input patch. Similarly, multi-scale and multi-resolution fusion are interchangeable here.)} 
    \label{fig:fig1_architecture}
\end{figure*}

Thus, we propose a new multi-resolution feature-level fusion based end-to-end crowd counting network to achieve the above objectives amid addressing the major crowd counting challenges. The proposed approach works at multiple scales via multi-columns, where each column primarily focuses on the respective scale (low-, medium-, or high-crowd), as shown in Fig. \ref{fig:fig1_architecture}. Unlike other state-of-the-art multi-scale or multi-column based methods, the columns also fuse and share the information with each other at a regular basis after every few deep layers (phase). Each column also takes the suitably rescaled version of the original input patch as its input prior without any classification process. Inspired by the success of high-resolution networks \cite{sun2019deep,wang2020deep}, each column also serves as a high-resolution sub-network, where the resolution is maintained the same as its input throughout the column. These repetitive multi-scale fusions, coupled with column-wise rescaled input priors and high-resolution maintenance, prove to be more effective in generalizing towards huge crowd variation issue (Objective \# 1) in comparison to recent state-of-the-art crowd counting methods as shown in the experiments section \ref{expi}. In addition, the simple yet effective column-wise input priors inclusion fulfills our objective \# 2 without using any compromising and extra crowd-density classification process. The contributions of this paper mainly include:

\begin{figure}[t]

	\begin{minipage}[b]{\columnwidth}
		\begin{center}
			\centerline{\includegraphics[width=0.995\columnwidth]{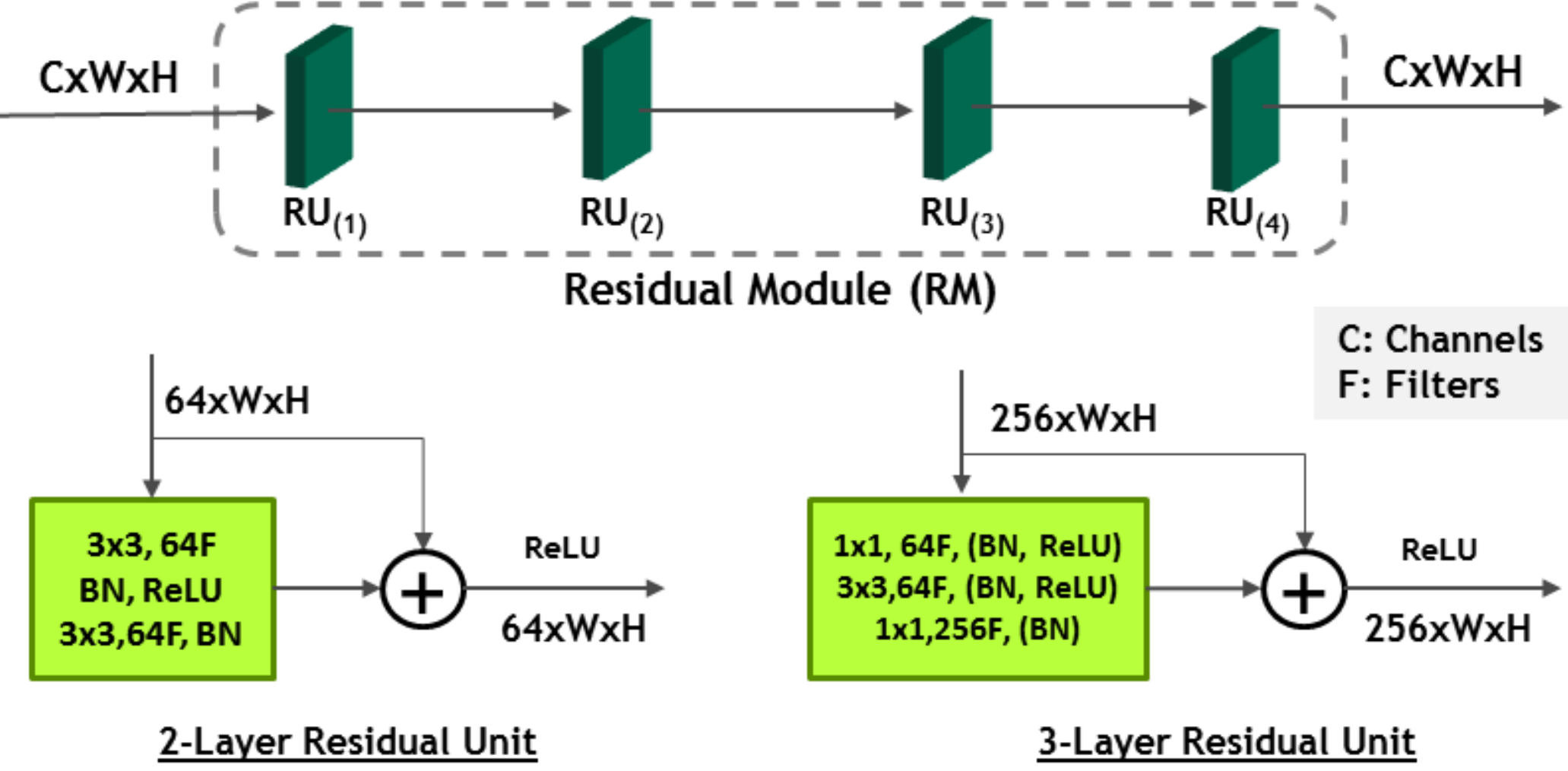}}
		\end{center}
	\end{minipage}		
	\caption{The Residual Module (RM) consists of either only 2- or 3-layers \cite{he2016deep} based four residual units (RU).}
    \label{fig:fig2_rm_and_rus}
\end{figure}

\begin{itemize}\setlength\itemsep{0.3em}
  \item We propose a new multi-resolution feature-level fusion based end-to-end crowd counting approach for still images that effectively deals with significant variations of crowd-density, lighting conditions, and large perspective.
  
  \item We propose an alternative patch rescaling module by more effectively using the input priors. Unlike the PRM \cite{sajid2020plug}, the proposed module fully utilizes all three crowd-density levels without requiring any compromising or additional crowd-density classification process.
  
  \item Quantitative experiments demonstrate that the proposed approach outperforms the state-of-the-art methods, including the PRM based schemes, by a large margin with up to 10\% improvements.
\\  
\end{itemize}

\section{Related Work}
Crowd estimation comes up with many key issues including huge crowd-density variation in and across different images, different illumination conditions, large perspective, and severe occlusions. Classical methods belong to either the detection-then-counting or regression-based schemes. Detection-based methods \cite{ge2009marked,li2008estimating,wang2011automatic,wu2005detection} were unable to work in case of high-dense crowd images, where it becomes really difficult to detect the handcrafted features. Regression-based methods \cite{chan2009bayesian,chen2012feature,ryan2009crowd} learn a transformation function to regress the crowd estimate from the local crowd features. These schemes also prove to be unreliable and ineffective due to the lack of generalization ability.

Recently, CNN based models are widely used due to their superior performance. Broadly, they belong to one of three types: Detection-based, regression-based, and density-map estimation methods. Detection-based methods \cite{shami2018people,li2019headnet} follow the principle of detection-then-counting, and use advance CNN detectors (e.g. Faster-RCNN \cite{girshick2015fast}, YOLO \cite{redmon2016you}) to detect persons in the images. Li \textit{et al.} \cite{li2019headnet} used the contextual information based adaptive head detection method for crowd count. Shami \textit{et al.} \cite{shami2018people} first detected persons using the CNN based head detectors, followed by the weighted average and final crowd count estimation. These methods seem impractical for high-dense crowd images due to the small head or person size. Regression-based methods learn a transformation function to map the input image to its crowd count. Wang \textit{et al.} \cite{wang2015deep} deployed the AlexNet \cite{krizhevsky2012imagenet} based architecture to perform crowd estimation on the input image. Fu \textit{et al.} \cite{fu2015fast} first classified the input 5-way based on the crowd-level, and then used two cascaded CNNs, where one improves the weaker crowd estimation being made by the other CNN as a boosting strategy. These methods alone fail to comprehend the hugely varying crowd-density scale. Sajid \textit{et al.} \cite{sajid2020zoomcount,sajid2020plug} proposed regression-based methods that use deep networks and smartly and accordingly rescaled input to estimate the crowd count. But the rescaling process first requires the expensive crowd-density classification process that comes with its own inaccuracies.

Density-map estimation based methods \cite{zhang2016single,cascadedmtl,sam2017switching,ranjan2018iterative,liu2018decidenet,wan2019residual,xu2019learn} generate crowd density-maps, with density value per pixel, and the final image crowd count is obtained by the summation of all pixels density estimations. Most recent state-of-the-art methods are the members of this category. Zhang \textit{et al.} \cite{zhang2016single} proposed a multi-column crowd counting network (MCNN), that uses three columns with different filter sizes to account for the respective crowd scale. Sindagi \textit{et al.} \cite{cascadedmtl} designed a cascaded end-to-end network that simultaneously calculates the crowd-density 10-way for the input and uses this classification as input prior to the next part of the network. Switch-CNN \cite{sam2017switching} uses a CNN-based switch to route the input patch to one of three specialized crowd regressors based on the crowd-density level. Ranjan \textit{et al.} \cite{ranjan2018iterative} designed the two-branch network, where the low-resolution branch have been combined with a high-resolution branch to generate the final density-map. Liu \textit{et al.} \cite{liu2018decidenet} proposed a hybrid approach that coupled both detection and density-map estimation techniques, and used the appropriate counting mode based on the crowd-density. Recently, Wan \textit{et al.} \cite{wan2019residual} used support image density-map to predict the input image density-map by the residual regression based difference between the two density-maps. Xu \textit{et al.} \cite{xu2019learn} first grouped patch-level density-maps into several density levels, followed by the automatic normalization via an online learning strategy with a multipolar center loss. One major issue with these methods is to find the optimal Gaussian kernel size, which depends on many related factors. They also do not generalize well on the huge crowd-variation challenge.

Thus, we propose a new multi-resolution feature-level fusion based end-to-end crowd counting network aiming to address the major crowd counting challenges and recent state-of-the-arts limitations.

\section{Proposed Approach}
The paper proposes a multi-column and multi-resolution fusion based end-to-end crowd counting network to achieve the two set objectives in Sec. \ref{intro}, amid addressing the major crowd counting challenges including huge crowd variation in and across different images, large perspective, and severe occlusions. The proposed scheme is shown in Fig. \ref{fig:fig1_architecture}, where the input image is first divided into $128 \times 128$ non-overlapping patches. Each resultant patch then goes through the proposed network for the patch-wise crowd count. Finally, the image crowd estimate is computed by the sum of the crowd count of all patches. The $128 \times 128$ input patch is used to generate the new $256 \times 256$ and $64 \times 64$ size input priors by $2\times$ times up- and down-scaling, respectively. These multi-scale input priors pass through the respective stems ($Stem_1, Stem_2, Stem_3$) to generate three separate initial channels ($IC_1, IC_2, IC_3$), which act as the corresponding input to three columns/branches in the main network (MN). The MN regularly fuses feature maps in between these branches. At the end of the main network, the resultant feature maps from three branches pass through the final regression head ($RH_{final}$) to yield the input patch crowd estimate. The MN also outputs into three auxiliary crowd estimating regression heads ($RH_1$, $RH_2$, $RH_3$) that helps in improving the input patch final crowd count. In the following, we will discuss three main components in detail. 

\setlength{\tabcolsep}{2.0pt}
\begin{table}[t]\small
\caption{\footnotesize Configurations of the Stems. Each conv operation denotes the Convolution-BN-ReLU series.}
	\begin{center}
	\begin{tabular}{|c|c|c|}
    \hline
 Name & Output size & Filters (F) Operation\\ \hline
\multicolumn{3}{|c|}{$Stem_1$} \\ \hline
 $I_1$ & $3 \times 256 \times 256$ &     \\ \hline
 & $64 \times 128 \times 128$ &  ($3 \times 3$) conv, stride 2, padding 1, $64$F    \\ \hline
  & $64 \times 64 \times 64$ &  ($3 \times 3$) conv, stride 2, padding 1, $64$F    \\ \hline
 $IC_1$ & $256 \times 64 \times 64$ &  ($1 \times 1$) conv, stride 1, padding 0, $256$F    \\ \hline
 \multicolumn{3}{|c|}{$Stem_2$} \\ \hline
 $I_2$ & $3 \times 128 \times 128$ &     \\ \hline
 & $64 \times 64 \times 64$ &    ($3 \times 3$) conv, stride 2, padding 1, $64$F    \\ \hline
$IC_2$ & $64 \times 32 \times 32$ &   ($3 \times 3$) conv, stride 2, padding 1, $64$F    \\ \hline
\multicolumn{3}{|c|}{$Stem_3$} \\ \hline
 $I_3$ & $3 \times 64 \times 64$ &     \\ \hline
 & $64 \times 32 \times 32$ &  ($3 \times 3$) conv, stride 2, padding 1, $64$F    \\ \hline
 $IC_3$ & $128 \times 16 \times 16$ & ($3 \times 3$) conv, stride 2, padding 1, $128$F    \\ \hline
	\end{tabular}
	\end{center}
	
	\label{table:config_cc1p}
\end{table}

\subsection{Input priors and respective stems}
We up- and down-scale the original $128 \times 128$ size input patch ($I_2$ or $P$) by $2\times$ to generate its rescaled versions ($256 \times 256$ and $64 \times 64$ respectively). These input priors ($I_1$, $I_2$, $I_3$) pass through their respective stems ($Stem_1$, $Stem_2$, $Stem_3$) to produce initial feature channels ($IC_1$, $IC_2$, $IC_3$). These stems, as shown in Table \ref{table:config_cc1p}, also decrease the input priors resolution to $1/4$, and the resultant initial feature maps resolution becomes half in the subsequent lower column. The upscaled input prior ($I_1$) helps in handling highly dense crowd regions by zooming in and observing the original input ($I_2$) in detail to avoid huge crowd under-estimation. Similarly, the input prior ($I_3$) uses a smaller scale, especially helpful for the low-crowd regions in the images that may otherwise cause significant crowd over-estimation. Empirically, it has been observed that coupling these simple yet effective rescaled input priors ($I_1$, $I_3$) with the original input ($I_2$) yields better crowd estimates, and consequently avoid huge crowd under- or over-estimation, as shown in the ablation study in Sec. \ref{ablation_input_priors}.

\setlength{\tabcolsep}{2.0pt}
\begin{table}[t]\small


	\caption{\footnotesize Standalone single-column output based $RH_{final}$ head versions (v1, v2, v3) and Auxiliary Crowd Regression Heads ($RH_1,RH_2,RH3$) configurations. Each conv operation denotes the Convolution-BN-ReLU sequence. These configurations mainly consist of several conv layers followed by the global average pooling and one or more fully connected (FC) layers to finally yield the crowd estimate (single neuron).}
	\vspace{-4mm}
	\begin{center}
	\begin{tabular}{|c|c|}
    \hline
  Output Size & Filters (F) Operation\\ \hline
\multicolumn{2}{|c|}{v1 (Highest-resolution)} \\ \hline
 $32 \times 64 \times 64$ &     \\ \hline
 $64 \times 32 \times 32$ &  ($3 \times 3$) conv, stride 2, padding 1, $64$F    \\ \hline
  $64 \times 16 \times 16$ &  ($3 \times 3$) conv, stride 2, padding 1, $64$F    \\ \hline
 $64 \times 8 \times 8$ &  ($2 \times 2$) Avg Pooling, stride 2    \\ \hline
 1024D, FC &  -    \\ \hline
 1D, FC (single neuron) &  -    \\ \hline
 \multicolumn{2}{|c|}{v2 (Middle-column)} \\ \hline
 $64 \times 32 \times 32$ &     \\ \hline
 $64 \times 32 \times 32$ &    ($1 \times 1$) conv, stride 1, padding 0, $64$F    \\ \hline
&   Rest continues as in v1 above    \\ \hline

\multicolumn{2}{|c|}{v3 (Lowest-resolution)} \\ \hline
 $128 \times 16 \times 16$ &     \\ \hline
 $64 \times 16 \times 16$ &    ($1 \times 1$) conv, stride 1, padding 0, $64$F    \\ \hline
 &   Rest continues as in v1 above    \\ \hline \hline
 

\multicolumn{2}{|c|}{$RH_1$ Configuration} \\ \hline
 $64 \times 32 \times 32$ &     \\ \hline
 $64 \times 16 \times 16$ &    ($3 \times 3$) conv, stride 2, padding 1, $64$F    \\ \hline
 & Rest continues as in v1 above    \\ \hline
 
\multicolumn{2}{|c|}{$RH_2$ and $RH_3$ Configuration} \\ \hline
 $128 \times 16 \times 16$ &     \\ \hline
 $64 \times 8 \times 8$ &    ($3 \times 3$) conv, stride 2, padding 1, $64$F    \\ \hline
 &   Rest continues as in v1 above    \\ \hline


	\end{tabular}
	\end{center}
	   
	\label{table:config_rhfinal_head}
    \vspace{-3mm}
\end{table}

\begin{figure}[t]
	\begin{minipage}[b]{1.0\columnwidth}
		\begin{center}
			\centerline{\includegraphics[width=1.0\columnwidth]{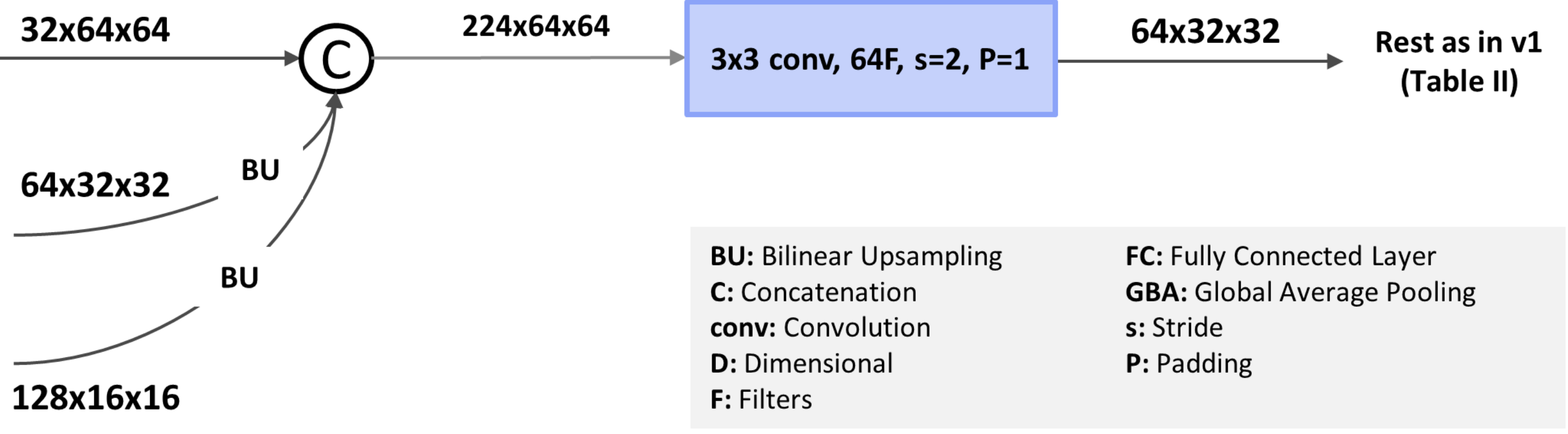}}
			\centerline{\footnotesize{(a) Concatenation-based (v4)}}
		\end{center}
	\end{minipage}
	\begin{minipage}[b]{1.0\columnwidth}
		\begin{center}
			\centerline{\includegraphics[width=1.0\columnwidth]{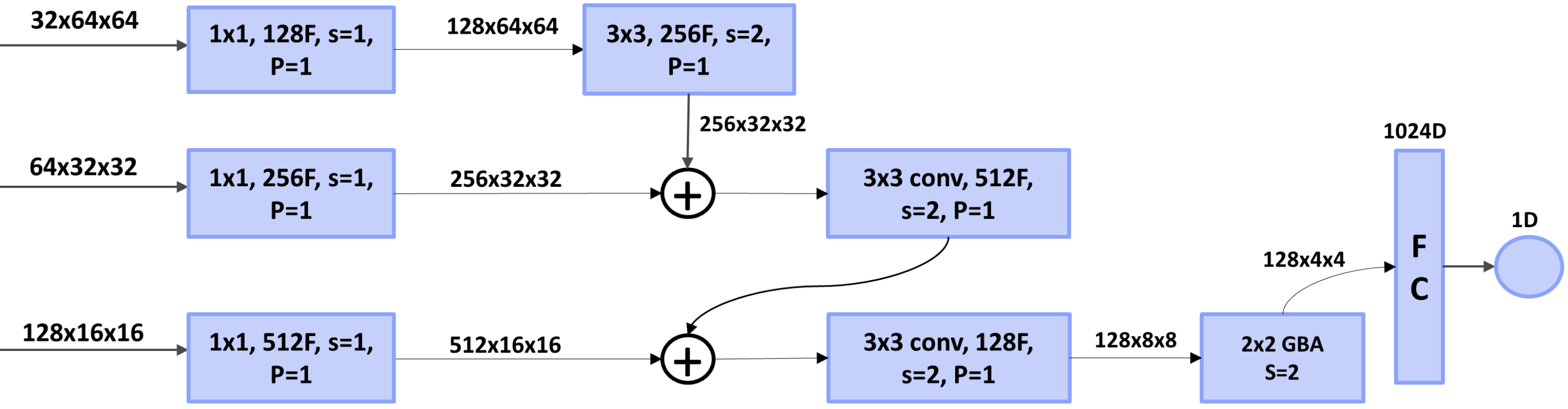}}
			\centerline{\footnotesize{(b) Summation-based (v5)}}
		\end{center}
	\end{minipage}
	\caption{\footnotesize{Concatenation-based crowd regression head (v4) concatenates the lower-resolutions with the highest-level channels using the bilinear upsampling, whereas the summation-based head (v5) adds the higher-level channels into the lowest-resolution feature maps, before proceeding through the several deep layers to finally yield the crowd estimate ($cc_{final}$) \cite{sun2019deep,wang2020deep}. 
	}}
    \label{fig:final_head_config123}
\end{figure}

\subsection{Main Network (MN)}
The main network is composed of three deep columns/branches, each with its own input prior feature maps, and also caters the respective crowd-density scale. The main network is divided into three phases from left to right, where each phase consists of one or more columns/branches. The total number of columns in a phase is equal to its phase number. All branches in a phase fuse feature maps with each other after each Residual Module (RM). At the end of each phase, the MN also feeds its lowest-resolution output into the auxiliary crowd regression heads ($RH_1, RH_2, RH_3$), as detailed in the next subsection \ref{crowd_reg_heads}. Each branch in the main network maintains its original input resolution throughout the branch, unlike other state-of-the-art multi-scale crowd estimation methods. The lower columns resolution and total channels in any phase depend on the highest-resolution branch ($i=1$). Let $C_1$ and $R_1$ be the total channels and their resolution respectively in the highest-resolution column. Then, the remaining columns ($i=2$,$3$) follow the below principle for their $C_i$ and $R_i$ in a given phase \cite{sun2019deep,wang2020deep}.

\vspace{-1mm}
\begin{equation}
\label{eq1}
C_i = 2C_{i-1} , R_i = \frac{R_{i-1}}{2}
\end{equation}

\textbf{Residual Module:} It consists of four residual units, where each unit is formed by either only 2-layer or 3-layer based residual block \cite{he2016deep}, as shown in Fig. \ref{fig:fig2_rm_and_rus}. The 2-layer based residual block \cite{he2016deep} contains two $3 \times 3$ convolution layers. Similarly, the 3-layer residual block \cite{he2016deep} starts with a bottleneck layer, followed by one $3 \times 3$ convolution layer and a bottleneck layer. Each convolution operation in these units is followed by the batch Normalization (BN) \cite{ioffe2015batch} and the nonlinear ReLU \cite{nair2010rectified} activation. Phase-1 uses the 2-layer based residual unit, whereas Phase-2 and 3 deploy the 3-layer based residual unit. The number of residual modules in each column per phase serves as a hyperparameter and discussed in ablation study in Sec. \ref{ablation_RM_quantity}. Moreover, by the network design, total residual modules in each column of a specific phase remain the same.\\

\textbf{Recurring Multi-resolution Fusions:} The primary purpose of the multi-resolution fusion is to exchange the information between different resolutions/columns, so as to enhance the generalization ability of the proposed scheme towards huge crowd diversity in and across different images. We utilize one or more $3 \times 3$ convolution operations to fuse higher-resolution feature maps into the lower-level channels. To fuse the lower-resolution feature maps into the higher-level channels, bilinear upsampling followed by the bottleneck layer (to adjust the number of channels) have been deployed. Let $Ch_i$ be the fusion source channels from column at $i$th index ($i=1,2$ or $3$), $Ch_j$ be the fusion target column at index $j$ ($j=1,2$ or $3$), and $f(.)$ be the transformation function. If $i<j$, then $f(Ch_i)$ downsamples the $Ch_i$ channels by $2(j-i)$ times via $(j-1)$ stride-2 $3 \times 3$ convolution(s). For example, fusing column-1 channels ($Ch_1$) into column-2 channels ($Ch_2$) first requires one stride-2 $3 \times 3$ convolution ($f(Ch_1)$) for $2\times$ downsampling. Similarly, $Ch_1$ fusion into $Ch_3$ requires 2 stride-2 $3 \times 3$ convolutions for $4\times$ downsampling before the fusion operation. If $i=j$, then $f(Ch_i)=Ch_i$, i.e., no transformation is done. If $i>j$, then $f(Ch_i)$ transformation upscales the $Ch_i$ using the bilinear upsampling, followed by the bottleneck layer to adjust the number of channels accordingly before the fusion process. Each convolution operation is followed by the Batch Normalization (BN) \cite{ioffe2015batch} and the nonlinear ReLU activation \cite{nair2010rectified}. After applying the appropriate transformation(s) and channels alignment(s) as discussed above, the summation based fusion operation finally outputs the sum of these transformed representations.

\subsection{Crowd Regression Heads}
\label{crowd_reg_heads}
The proposed approach contains three phase-wise crowd regression heads ($RH_1,RH_2,RH_3$) and the final regression head ($RH_{final}$).

\textbf{Phase-wise Regression Heads:} One of the primary purpose of phase based organization of the main network is to introduce auxiliary crowd regression heads ($RH_1,RH_2,RH_3$) at the end of each phase. The last lowest-resolution output of each phase serves as the input to its respective regression head. These heads mainly consist of several convolution based deep layers, followed by optional average pooling operation and one or more fully connected (FC) layers as detailed in Table \ref{table:config_rhfinal_head}. Finally, the single neuron ($1D,FC$) at the end of each head gives the corresponding crowd counts ($cc_{P(1)},cc_{P(2)},cc_{P(3)}$) for the input patch ($P$).

\textbf{Final Regression Head ($RH_{final}$):} Phase-3 outputs three blocks of feature maps, each from the respective column with varying resolution. These blocks have been exploited in different ways for possible and effective $RH_{final}$ head configuration, as discussed below.

\textit{Standalone Single-Column Output based (v1,v2,v3).} Here, we only use one of three phase-3 outputs for the $RH_{final}$ configuration \cite{sun2019deep,wang2020deep}. Subsequent configurations are shown in Table \ref{table:config_rhfinal_head}, and named as v1 (highest-resolution), v2 (middle-column), and v3 (lowest-resolution), respectively. These representations consist of several deep layers, followed by the 1024 dimensional fully connected (FC) layer and the final single neuron to directly regress the crowd count.

\textit{Concatenation-based (v4).} The lower-resolution feature maps concatenate at the highest-resolution branch, with configuration shown in Fig. \ref{fig:final_head_config123}(a) \cite{sun2019deep,wang2020deep}.

\textit{Summation-based (v5).} The higher-level feature maps are summed up into the subsequent lower resolution feature maps after respective downscaling, as shown in Fig. \ref{fig:final_head_config123}(b) \cite{sun2019deep,wang2020deep}.

Employing one of the above configurations, the $RH_{final}$ yields its crowd count ($cc_{final}$) for the input patch $P$. The final crowd count ($CC_P$) for the original input patch $P$ is computed using all regression heads weighted crowd estimates as follows:

\vspace{-2mm}
\begin{equation}
\label{eq2}
CC_P = w*cc_{P(1)} + x*cc_{P(2)} + y*cc_{P(3)} + z*cc_{final}
\end{equation}

\noindent Where $w=x=y=0.1$ and $z=0.7$. The mean squared error (MSE) has been used as the loss function for each of the four regression heads (RH), given as follows: 

\vspace{-2mm}
\begin{equation}
\label{eq3}
L_{RH} = \frac{1}{N} \sum_{i=1}^{N} (F(x_i,\Theta)-y_i)^{2}
\end{equation}
where $N$ represents the total training patches per batch, $y_i$ denotes the ground truth crowd count for the input image patch $x_i$, and $F(.)$ represents the transformation function that learns the $x_i$ to crowd count mapping with learnable weights $\Theta$. Finally, the total loss for the input patch $P$ is the weighted accumulation of all four regression head losses as below:

\vspace{-2mm}
\begin{equation}
\label{eq4}
L_P = w*L_{RH_1} + x*L_{RH_2} + y*L_{RH_3} + z*L_{final}
\end{equation}

\section{Implementation Details}
We employ the following two standard metrics, namely Mean Absolute Error (MAE) and Root Mean Square Error (RMSE), for the evaluation and comparison of the proposed scheme with other state-of-the-art methods.

\vspace{-1mm}
\begin{equation}
\label{eq3}
MAE = \frac{1}{T} \sum_{t=1}^{T} |CC_{t}-\hat{CC_{t}}|, RMSE =\sqrt[]{ \frac{1}{T} \sum_{t=1}^{T} (CC_{t}-\hat{CC_{t}})^{2}}
\end{equation}
where $T$ represents the total test images in a dataset, and $CC_t$ and $\hat{CC_{t}}$ denote the actual and estimated crowd counts respectively for the test image $t$.

\textbf{Training Details:} We randomly extract $60,000$ patches of $256 \times 256$, $128 \times 128$, and $64 \times 64$ sizes with varying crowd number from the training images. Horizontal flip based data augmentation is then used to double the training samples quantity. We trained the proposed model for $100$ epochs, used SGD optimizer with a weight decay of $0.0001$ and a Nesterov momentum value of $0.9$. Multi-step learning has been employed that initially starts at $0.001$ and decreases by half after every 25 epochs. As per the standard literature convention, $10\%$ data from the predefined training set has been separated for the model validation purpose.

\setlength{\tabcolsep}{2.0pt}
\begin{table}[t]\small

	\caption{\footnotesize Experiments on ShanghaiTech \cite{zhang2016single} and UCF-QNRF \cite{idrees2018composition} benchmarks. The proposed method (v5) outperforms the state-of-the-art methods (including the PRM based approach \cite{sajid2020plug}) for the RMSE metric, while giving comparable results for the MAE metric. Other versions of the proposed scheme (v1, v2, v3, v4) also  perform effectively well.}

	\begin{center}
	\begin{tabular}{|c|c|c|c|c|}
    \hline

 & \multicolumn{2}{c|}{ShanghaiTech} & \multicolumn{2}{c|}{UCF-QNRF}\\ \hline
Method & MAE  & RMSE & MAE  & RMSE\\ \hline
MCNN \cite{zhang2016single}  & 110.2  & 173.2 & 277   & 426   \\ \hline
CMTL \cite{cascadedmtl} & 101.3  & 152.4 & 252   & 514   \\ \hline
Switch-CNN \cite{sam2017switching} & 90.4 & 135.0 & 228   & 445   \\ \hline
SaCNN \cite{zhang2018crowd} & 86.8 & 139.2 & -   & -   \\ \hline
IG-CNN \cite{babu2018divide} & 72.5 & 118.2  & -   & -  \\ \hline
ACSCP \cite{shen2018crowd}   & 75.7  & 102.7 & -   & - \\ \hline 
CSRNet \cite{li2018csrnet}   & 68.2  & 115.0 & -   & - \\ \hline 

CL\cite{idrees2018composition} & - & - & 132 & 191    \\ \hline

CFF \cite{shi2019counting} & 65.2 & 109.4 & 93.8   & 146.5 \\ \hline
RRSP \cite{wan2019residual} & 63.1 & 96.2  & -   & -  \\ \hline 
CAN \cite{liu2019context} & \textbf{62.3} & 100.0 & 107   & 183   \\ \hline
L2SM \cite{xu2019learn} & 64.2 & 98.4 & 104.7   & 173.6 \\ \hline

BL \cite{ma2019bayesian} & 62.8 & 101.8 & \textbf{88.7} & 154.8    \\ \hline

ZoomCount \cite{sajid2020zoomcount} & 66.6 & 94.5  & 128 & 201 \\ \hline
PRM-based\cite{sajid2020plug} & 67.8 & 86.2  & 94.5 & 141.9 \\ \hline \hline

v1/v2 (\textbf{ours}) & 71.4/70.1 & 85.7/85.3 & 103.1/100.6 & 139.6/136.3   \\ \hline
v3/v4 (\textbf{ours}) & 69.8/67.9 & 84.7/81.9 & 101.7/98.4 & 137/135.1   \\ \hline
\textbf{v5} (\textbf{ours})  & 67.1  &  \textbf{81.0} & 96.9  & \textbf{130.1}   \\ \hline

	\end{tabular}
	\end{center}
	   
	\label{table:ST_results}
\end{table}

\setlength{\tabcolsep}{2.0pt}
\begin{table}[t]\small

	\caption{\footnotesize AHU-Crowd dataset experiments. The proposed approach outperforms the state-of-the-arts for both metrics.}
	\begin{center}
	\begin{tabular}{|c|c|c|c|}
    \hline
Method & MAE & RMSE\\ \hline
Haar Wavelet \cite{oren1997pedestrian} & 409.0   & -    \\ \hline
DPM \cite{felzenszwalb2008discriminatively} & 395.4  & -    \\ \hline
BOW–SVM \cite{csurka2004visual} & 218.8  & -    \\ \hline
Ridge Regression \cite{chen2012feature} & 207.4  & -    \\ \hline
Hu et al. \cite{hu2016dense} & 137  & -    \\ \hline
DSRM \cite{yao2017deep} &  81  &  129    \\ \hline
ZoomCount \cite{sajid2020zoomcount} & 74.9 & 111 \\ \hline
CC-2P (PRM-based)\cite{sajid2020plug} & 66.6 & 101.9\\ \hline \hline

v1/v2/v3 (\textbf{ours}) & 69.8/67.1/65.4 & 107.8/103.5/100.2   \\ \hline
v4/\textbf{v5} (\textbf{ours}) & 63.1/\textbf{60.2} & 99.5/\textbf{91.7} \\ \hline

	\end{tabular}
    \vspace{0mm}
	\end{center}

	\label{table:AHU_results}
\end{table}

\begin{figure*}
	\begin{minipage}[b][][b]{0.4\columnwidth}
		\begin{center}
			\centerline{\includegraphics[width=1\columnwidth]{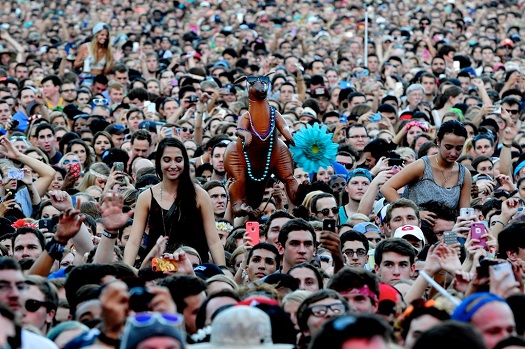}}
			\centerline{\footnotesize{GT=597, PRM=431}}
			\centerline{\footnotesize{Ours=595, DME=301}}
		\end{center}
	\end{minipage}
		\begin{minipage}[b][][b]{0.4\columnwidth}
		\begin{center}
			\centerline{\includegraphics[width=1\columnwidth]{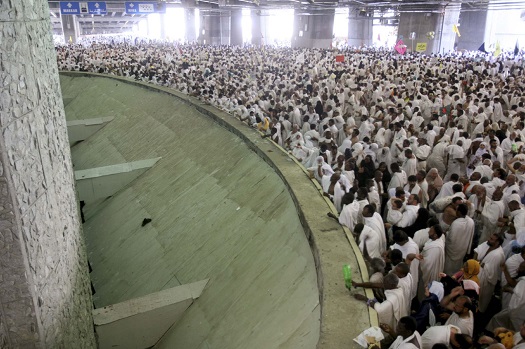}}
			\centerline{\footnotesize{GT=1929, PRM=1395}}
			\centerline{\footnotesize{Ours=1920, DME=623}}
		\end{center}
	\end{minipage}
	\begin{minipage}[b][][b]{0.4\columnwidth}
		\begin{center}
			\centerline{\includegraphics[width=1\columnwidth]{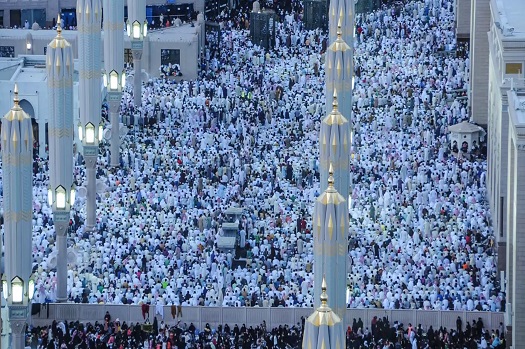}}
			\centerline{\footnotesize{GT=3653, PRM=2792}}
			\centerline{\footnotesize{Ours=3639, DME=2792}}
		\end{center}
	\end{minipage}
	\begin{minipage}[b][][b]{0.4\columnwidth}
		\begin{center}
			\centerline{\includegraphics[width=1\columnwidth]{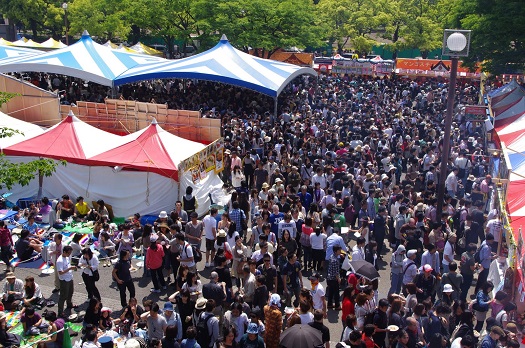}}
			\centerline{\footnotesize{GT=1070, PRM=1011}}
			\centerline{\footnotesize{Ours=1072, DME=722}}
		\end{center}
	\end{minipage}			
		\begin{minipage}[b][][b]{0.4\columnwidth}
		\begin{center}
			\centerline{\includegraphics[width=1\columnwidth]{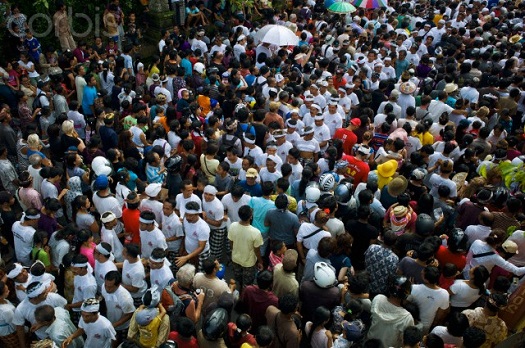}}
			\centerline{\footnotesize{GT=518, PRM=417}}
			\centerline{\footnotesize{Ours=509, DME=293}}
		\end{center}
	\end{minipage}	
					
    \vspace{-6mm}
	\caption{\footnotesize{Ground truth (GT) based qualitative comparison.
	}}
	\label{fig:qualityResults}
\end{figure*}

\setlength{\tabcolsep}{2.0pt}
\begin{table}[t]\small
\vspace{-2mm}
\caption{\footnotesize Three respective ablation studies on the effect of RM modules quantity per column in Phase-2 and 3, Input priors ($I_1,I_2,I_3$), and suxiliary crowd regression heads usage on the proposed network performance. The results demonstrate the fact that the priors and auxiliary heads are of vital importance, as the MAE and RMSE errors increase without them. These ablation experiments are done using the ShanghaiTech dataset, and performed on the proposed method version v5, as being the best of them quantitatively.}

	\begin{center}
	\begin{tabular}{|c|c|c|c|c|}
    \hline
    
 \multicolumn{3}{|c|}{RM Modules Quantity Effect}\\ \hline
 
  RM Modules per column in Phase-2 and 3 & MAE  & RMSE\\ \hline
 1 & 79.3 & 111.4   \\ \hline
 \textbf{2} (our default) & \textbf{67.1}  & \textbf{81.0}    \\ \hline
 3 & 75.8 & 104.7   \\ \hline
 \hline
 \multicolumn{3}{|c|}{Input Priors Effect}\\ \hline
 
  & MAE  & RMSE\\ \hline
 w/o ($I_2,I_3$) & 77.1  & 108.8    \\ \hline
 w/o ($I_3$) & 75.9 & 106.5   \\ \hline
 w/o ($I_2$) & 73.8 & 101.4   \\ \hline
 only ($I_1$) with original input size ($256\times256$) & 80.1 & 124.5   \\ \hline
 with ($I_1,I_2,I_3$) (our default) & \textbf{67.1} & \textbf{81.0} \\ \hline
\hline
\multicolumn{3}{|c|}{Auxiliary Regression Heads Effect}\\ \hline
 
   & MAE  & RMSE\\ \hline
 w/o $RH_1$ & 76.2 & 107.0    \\ \hline
 w/o $RH_2$ & 71.7 & 115.2    \\ \hline
 w/o $RH_3$ & 73.9 & 103.1    \\ \hline
 w/o ($RH_1,RH_2,RH_3$) & 78.5 & 120.7    \\ \hline
 with ($RH_1,RH_2,RH_3$) (our default) & \textbf{67.1} & \textbf{81.0} \\ \hline
 
	\end{tabular}
	\end{center}
	
	\label{table:abl_results}
    \vspace{-2mm}
\end{table}

\setlength{\tabcolsep}{1.0pt}
\begin{table}\small 
\caption{\footnotesize Cross-dataset experiments demonstrate the better generalization capability of the proposed approach.}
\vspace{-2mm}
	\begin{center}
	\begin{tabular}{|c|c|c|}
    \hline
Method & MAE  & RMSE\\ \hline
Cascaded-mtl \cite{cascadedmtl} & 308  & 478  \\ \hline
Switch-CNN \cite{sam2017switching} & 301  & 457   \\ \hline
CC-2P (PRM based) \cite{sajid2020plug} & 219 & 305    \\ \hline 

v1/v2/v3 (\textbf{ours}) & 214/217/212 & 301/303/294    \\ \hline

v4/\textbf{v5} (\textbf{ours}) & 206/\textbf{201} & 285/\textbf{278}    \\ \hline

	\end{tabular}
	\end{center}
	
	\label{table:transferLearning}
    \vspace{-6mm}
\end{table}

\section{Experimental results}
\label{expi}
In this section, we first perform standard quantitative analysis on three benchmark datasets: UCF-QNRF \cite{idrees2018composition}, ShanghaiTech \cite{zhang2016single}, and AHU-crowd \cite{hu2016dense}. These benchmarks pose a great collective challenge for the proposed scheme to prove its effectiveness, as they vary significantly with each other in terms of average image resolution, average crowd number per image, total images, and lighting conditions. Next, we discuss the ablation experiments findings and the cross-dataset evaluation, followed by the qualitative evaluation. For comparison with other state-of-the-art methods, we evaluate all five versions of the proposed method ($v1, v2, v3, v4, v5$) as discussed in Sec. \ref{crowd_reg_heads}.

\subsection{Experiments on UCF-QNRF Dataset}
UCF-QNRF \cite{idrees2018composition} is one of the most diverse, realistic, and challenging dataset. It consists of $1,535$ free-view images with a predefined train/test division of $1,201/334$. It contains images with relatively very small ($300\times377$) and very large ($6666\times9999$) resolutions, with $1,251,642$ total people annotations that show its crowd complexity and diversity. We compare the proposed approach with the state-of-the-art methods (including the PRM based approach \cite{sajid2020plug}) in Table \ref{table:ST_results}. As shown, the proposed scheme (v5) outperforms the state-of-the-arts under the RMSE evaluation metric by $\sim8.3\%$ (from $141.9$ to $130.1$), amid performing reasonably well for the MAE. 

\subsection{Experiments on ShanghiTech Dataset}
The ShanghaiTech Part-A benchmark \cite{zhang2016single} is another diverse and free-view crowd counting benchmark. It contains $482$ images (predefined train/test division of $300/182$) with a total of $241,677$ people annotations and average image resolution of $589\times868$. Based on the quantitative comparison with the state-of-the-art methods (including the PRM based methods \cite{sajid2020plug}) as shown in Table \ref{table:ST_results}, the proposed approach (v5) decreases the RMSE error by $\sim6\%$ (from $86.2$ to $81.0$). For the MAE metric, our schemes give reasonable and comparable results. The lowest RMSE value also demonstrates that our method is less susceptible to huge crowd over- and under-estimation.

\subsection{Experiments on AHU-Crowd Dataset}
The AHU-Crowd \cite{hu2016dense} dataset contains $107$ crowd images with $58$ to $2,201$ people annotations per image and $45,807$ annotations in total. As per the standard evaluation process, we perform 5-fold cross-validation, and final (MAE, RMSE) results are obtained by computing their average. Evaluation and comparison results are shown in Table \ref{table:AHU_results}, where our scheme (v5) outperforms other state-of-the-arts under both evaluation metrics with significant improvements i.e., the MAE error decreases by $\sim9.6\%$ (from $66.6$ to $60.2$) and the RMSE improves by $\sim10\%$ (from $101.9$ to $91.7$).

\subsection{Effect of RM Modules Quantity}
\label{ablation_RM_quantity}
In this ablation study, we examine the effect of the number of RM modules in the Phase-2 and 3 of the proposed scheme. Instead of using 2 RM modules by default, we evaluate our method (v5) separately by utilizing only either 1 or 3 RM modules per column in each phase. As shown in Table \ref{table:abl_results} on the ShanghaiTech \cite{zhang2016single} dataset, our default choice of 2 RM modules per column in both phases (Phase-2 and 3) yields the most effective results. Using 1 or 3 RM modules per column in each phase cause the MAE, RMSE errors increase of ($15.4\%,27.3\%$) and ($11.5\%,22.6\%$) respectively. Thus, we have employed 2 RM modules per column in Phase-2 and 3.

\subsection{Effect of Input Priors ($I_1,I_2,I_3$)}
\label{ablation_input_priors}
This section reveals the quantitative importance of the input priors. We remove these input priors in different experimental settings to analyze their effectiveness. In the first three separate experiments, we only use ($I_1$), ($I_1,I_2$) and ($I_1,I_3$) input prior(s) respectively. While, in the last setting, we only deployed the $I_1$ input, but with the original $256\times256$ input size without any rescaling. The consequent ablation results are shown in Table \ref{table:abl_results}, from which we can see that removing these input priors significantly decreases the overall network performance (with minimum MAE, RMSE errors increase of $9.1\%,20.1\%$ respectively). Thus, all three input priors are critical for the proposed method effectiveness. 

\subsection{Effect of Auxiliary Crowd Regression Heads}
\label{ablation_aux_reg_heads}
In this experiment, we analyze the quantitative effect of employing the auxiliary crowd regression heads ($RH_1, RH_2, RH_3$) in the proposed scheme. During this ablation study, we removed each auxiliary head one by one and evaluate the network (v5) on the ShanghaiTech Part-A \cite{zhang2016single} dataset. As shown in Table \ref{table:abl_results}, the performance decreases significantly after removing these heads ($RH_1,RH_2,RH_3$). For instance, without using the $RH_1$ head, the MAE error increases the most with a jump of $11.9\%$. Similarly, the RMSE error is being affected the most by the $RH_2$ head removal with a $29.7\%$ increase in error.

\subsection{Cross-Dataset Evaluation}
To analyze the generalization ability of the proposed method, we carried out the cross-dataset validation. During the experiment, all methods have been trained and tested on the ShanghaiTech Part-A \cite{zhang2016single} and the UCF-QNRF \cite{idrees2018composition} datasets respectively. As shown in Table \ref{table:transferLearning}, the proposed method demonstrates better generalization capability as compared to the state-of-the-art methods (including the PRM-based scheme \cite{sajid2020plug}) with MAE, RMSE errors decrease by $8.2\%$ (from $219$ to $201$) and $8.9\%$ (from $305$ to $278$) respectively. Similar to the previous experiments, the proposed approach version (v5) appears to be the most effective cross-dataset validation scheme with the lowest MAE, RMSE values.

\subsection{Qualitative Evaluation}
In this section, we demonstrate some qualitative results as shown in Fig. \ref{fig:qualityResults}. We also compare our scheme with the PRM-based \cite{sajid2020plug} and density-map estimation (DME) \cite{idrees2018composition} based recent state-of-the-art methods. In comparison, it can be observed that the proposed scheme yields the best performance of all on these actual test images with hugely varying crowd-density, lighting condition, and image resolution.

\section{Conclusion}
To address the major crowd count challenges, we proposed a new multi-resolution fusion based end-to-end crowd counting network for the still images in this work. We also deployed a new and effective PRM substitute that uses three input priors, and proves to be much more accurate than the PRM. Both quantitative and qualitative results have revealed that the proposed network outperforms the state-of-the-art approaches under the RMSE evaluation metric. Cross-dataset evaluation also demonstrates better generalization capability of our approach towards new datasets.

\balance
\bibliographystyle{IEEEtranS}
\bibliography{egbib}

\begin{thebibliography}{10}
\providecommand{\url}[1]{#1}
\csname url@samestyle\endcsname
\providecommand{\newblock}{\relax}
\providecommand{\bibinfo}[2]{#2}
\providecommand{\BIBentrySTDinterwordspacing}{\spaceskip=0pt\relax}
\providecommand{\BIBentryALTinterwordstretchfactor}{4}
\providecommand{\BIBentryALTinterwordspacing}{\spaceskip=\fontdimen2\font plus
\BIBentryALTinterwordstretchfactor\fontdimen3\font minus
  \fontdimen4\font\relax}
\providecommand{\BIBforeignlanguage}[2]{{%
\expandafter\ifx\csname l@#1\endcsname\relax
\typeout{** WARNING: IEEEtranS.bst: No hyphenation pattern has been}%
\typeout{** loaded for the language `#1'. Using the pattern for}%
\typeout{** the default language instead.}%
\else
\language=\csname l@#1\endcsname
\fi
#2}}
\providecommand{\BIBdecl}{\relax}
\BIBdecl

\bibitem{babu2018divide}
D.~Babu~Sam, N.~N. Sajjan, R.~Venkatesh~Babu, and M.~Srinivasan, ``{D}ivide and
  {G}row: {C}apturing {H}uge {D}iversity in {C}rowd {I}mages {W}ith
  {I}ncrementally {G}rowing {CNN},'' in \emph{Proceedings of the IEEE
  Conference on Computer Vision and Pattern Recognition}, 2018, pp. 3618--3626.

\bibitem{cen2019dictionary}
F.~Cen and G.~Wang, ``Dictionary representation of deep features for
  occlusion-robust face recognition,'' \emph{IEEE Access}, vol.~7, pp.
  26\,595--26\,605, 2019.

\bibitem{chan2009bayesian}
A.~B. Chan and N.~Vasconcelos, ``Bayesian poisson regression for crowd
  counting,'' in \emph{IEEE International Conference on Computer Vision}, 2009,
  pp. 545--551.

\bibitem{chen2012feature}
K.~Chen, C.~C. Loy, S.~Gong, and T.~Xiang, ``Feature mining for localised crowd
  counting.'' in \emph{BMVC}, vol.~1, no.~2, 2012, p.~3.

\bibitem{csurka2004visual}
G.~Csurka, C.~Dance, L.~Fan, J.~Willamowski, and C.~Bray, ``Visual
  categorization with bags of keypoints,'' in \emph{Workshop on statistical
  learning in computer vision, ECCV}, vol.~1, no. 1-22, 2004, pp. 1--2.

\bibitem{felzenszwalb2008discriminatively}
P.~Felzenszwalb, D.~McAllester, and D.~Ramanan, ``A discriminatively trained,
  multiscale, deformable part model,'' in \emph{Computer Vision and Pattern
  Recognition, CVPR}, 2008, pp. 1--8.

\bibitem{fu2015fast}
M.~Fu, P.~Xu, X.~Li, Q.~Liu, M.~Ye, and C.~Zhu, ``Fast crowd density estimation
  with convolutional neural networks,'' \emph{Engineering Applications of
  Artificial Intelligence}, vol.~43, pp. 81--88, 2015.

\bibitem{ge2009marked}
W.~Ge and R.~T. Collins, ``Marked point processes for crowd counting,'' in
  \emph{2009 IEEE Conference on Computer Vision and Pattern Recognition}.\hskip
  1em plus 0.5em minus 0.4em\relax IEEE, 2009, pp. 2913--2920.

\bibitem{girshick2015fast}
R.~Girshick, ``Fast {R-CNN},'' in \emph{Proceedings of the IEEE international
  conference on computer vision}, 2015, pp. 1440--1448.

\bibitem{he2016deep}
K.~He, X.~Zhang, S.~Ren, and J.~Sun, ``Deep residual learning for image
  recognition,'' in \emph{Proceedings of the IEEE conference on computer vision
  and pattern recognition}, 2016, pp. 770--778.

\bibitem{hu2016dense}
Y.~Hu, H.~Chang, F.~Nian, Y.~Wang, and T.~Li, ``Dense crowd counting from still
  images with convolutional neural networks,'' \emph{Journal of Visual
  Communication and Image Representation}, vol.~38, pp. 530--539, 2016.

\bibitem{idrees2018composition}
H.~Idrees, M.~Tayyab, K.~Athrey, D.~Zhang, S.~Al-Maadeed, N.~Rajpoot, and
  M.~Shah, ``Composition loss for counting, density map estimation and
  localization in dense crowds,'' in \emph{Proceedings of the European
  Conference on Computer Vision (ECCV)}, 2018, pp. 532--546.

\bibitem{ioffe2015batch}
S.~Ioffe and C.~Szegedy, ``Batch normalization: Accelerating deep network
  training by reducing internal covariate shift,'' \emph{arXiv preprint
  arXiv:1502.03167}, 2015.

\bibitem{krizhevsky2012imagenet}
A.~Krizhevsky, I.~Sutskever, and G.~E. Hinton, ``Imagenet classification with
  deep convolutional neural networks,'' in \emph{Advances in neural information
  processing systems}, 2012, pp. 1097--1105.

\bibitem{li2008estimating}
M.~Li, Z.~Zhang, K.~Huang, and T.~Tan, ``Estimating the number of people in
  crowded scenes by mid based foreground segmentation and head-shoulder
  detection,'' in \emph{2008 19th International Conference on Pattern
  Recognition}.\hskip 1em plus 0.5em minus 0.4em\relax IEEE, 2008, pp. 1--4.

\bibitem{li2019headnet}
W.~Li, H.~Li, Q.~Wu, F.~Meng, L.~Xu, and K.~N. Ngan, ``Headnet: An end-to-end
  adaptive relational network for head detection,'' \emph{IEEE Transactions on
  Circuits and Systems for Video Technology}, 2019.

\bibitem{li2018csrnet}
Y.~Li, X.~Zhang, and D.~Chen, ``{CSRN}et: Dilated convolutional neural networks
  for understanding the highly congested scenes,'' in \emph{Proceedings of the
  IEEE Conference on Computer Vision and Pattern Recognition}, 2018, pp.
  1091--1100.

\bibitem{liu2018decidenet}
J.~Liu, C.~Gao, D.~Meng, and A.~G. Hauptmann, ``Decide{N}et: {c}ounting varying
  density crowds through attention guided detection and density estimation,''
  in \emph{Proceedings of the IEEE Conference on Computer Vision and Pattern
  Recognition}, 2018, pp. 5197--5206.

\bibitem{liu2019context}
W.~Liu, M.~Salzmann, and P.~Fua, ``Context-aware crowd counting,'' in
  \emph{Proceedings of the IEEE Conference on Computer Vision and Pattern
  Recognition}, 2019, pp. 5099--5108.

\bibitem{ma2020mdfn}
W.~Ma, Y.~Wu, F.~Cen, and G.~Wang, ``Mdfn: Multi-scale deep feature learning
  network for object detection,'' \emph{Pattern Recognition}, vol. 100, p.
  107149, 2020.

\bibitem{ma2019bayesian}
Z.~Ma, X.~Wei, X.~Hong, and Y.~Gong, ``Bayesian loss for crowd count estimation
  with point supervision,'' in \emph{Proceedings of the IEEE International
  Conference on Computer Vision}, 2019, pp. 6142--6151.

\bibitem{nair2010rectified}
V.~Nair and G.~E. Hinton, ``Rectified linear units improve restricted boltzmann
  machines,'' in \emph{ICML}, 2010.

\bibitem{oren1997pedestrian}
M.~Oren, C.~Papageorgiou, P.~Sinha, E.~Osuna, and T.~Poggio, ``Pedestrian
  detection using wavelet templates,'' in \emph{cvpr}, vol.~97, 1997, pp.
  193--199.

\bibitem{ranjan2018iterative}
V.~Ranjan, H.~Le, and M.~Hoai, ``Iterative crowd counting,'' in
  \emph{Proceedings of the European Conference on Computer Vision (ECCV)},
  2018, pp. 270--285.

\bibitem{redmon2016you}
J.~Redmon, S.~Divvala, R.~Girshick, and A.~Farhadi, ``You only look once:
  Unified, real-time object detection,'' in \emph{Proceedings of the IEEE
  conference on computer vision and pattern recognition}, 2016, pp. 779--788.

\bibitem{ryan2009crowd}
D.~Ryan, S.~Denman, C.~Fookes, and S.~Sridharan, ``Crowd counting using
  multiple local features,'' in \emph{Digital Image Computing: Techniques and
  Applications, DICTA}, 2009, pp. 81--88.

\bibitem{sajid2020zoomcount}
U.~Sajid, H.~Sajid, H.~Wang, and G.~Wang, ``Zoomcount: A zooming mechanism for
  crowd counting in static images,'' \emph{IEEE Transactions on Circuits and
  Systems for Video Technology}, 2020.

\bibitem{sajid2020plug}
U.~Sajid and G.~Wang, ``Plug-and-play rescaling based crowd counting in static
  images,'' in \emph{The IEEE Winter Conference on Applications of Computer
  Vision}, 2020, pp. 2287--2296.

\bibitem{sam2017switching}
D.~B. Sam, S.~Surya, and R.~V. Babu, ``Switching convolutional neural network
  for crowd counting,'' in \emph{2017 IEEE Conference on Computer Vision and
  Pattern Recognition (CVPR)}.\hskip 1em plus 0.5em minus 0.4em\relax IEEE,
  2017, pp. 4031--4039.

\bibitem{shami2018people}
M.~Shami, S.~Maqbool, H.~Sajid, Y.~Ayaz, and S.-C.~S. Cheung, ``People counting
  in dense crowd images using sparse head detections,'' \emph{IEEE Transactions
  on Circuits and Systems for Video Technology}, 2018.

\bibitem{shen2018crowd}
Z.~Shen, Y.~Xu, B.~Ni, M.~Wang, J.~Hu, and X.~Yang, ``Crowd counting via
  adversarial cross-scale consistency pursuit,'' in \emph{Proceedings of the
  IEEE conference on computer vision and pattern recognition}, 2018, pp.
  5245--5254.

\bibitem{shi2019counting}
Z.~Shi, P.~Mettes, and C.~G. Snoek, ``Counting with focus for free,''
  \emph{arXiv preprint arXiv:1903.12206}, 2019.

\bibitem{cascadedmtl}
V.~A. Sindagi and V.~M. Patel, ``{CNN}-based cascaded multi-task learning of
  high-level prior and density estimation for crowd counting,'' in \emph{IEEE
  International Conference on Advanced Video and Signal Based Surveillance
  (AVSS)}, 2017, pp. 1--6.

\bibitem{sun2019deep}
K.~Sun, B.~Xiao, D.~Liu, and J.~Wang, ``Deep high-resolution representation
  learning for human pose estimation,'' in \emph{Proceedings of the IEEE
  conference on computer vision and pattern recognition}, 2019, pp. 5693--5703.

\bibitem{wan2019residual}
J.~Wan, W.~Luo, B.~Wu, A.~B. Chan, and W.~Liu, ``Residual regression with
  semantic prior for crowd counting,'' in \emph{Proceedings of the IEEE
  Conference on Computer Vision and Pattern Recognition}, 2019, pp. 4036--4045.

\bibitem{wang2015deep}
C.~Wang, H.~Zhang, L.~Yang, S.~Liu, and X.~Cao, ``Deep people counting in
  extremely dense crowds,'' in \emph{Proceedings of the 23rd ACM international
  conference on Multimedia}, 2015, pp. 1299--1302.

\bibitem{wang2020deep}
J.~Wang, K.~Sun, T.~Cheng, B.~Jiang, C.~Deng, Y.~Zhao, D.~Liu, Y.~Mu, M.~Tan,
  X.~Wang \emph{et~al.}, ``Deep high-resolution representation learning for
  visual recognition,'' \emph{IEEE transactions on pattern analysis and machine
  intelligence}, 2020.

\bibitem{wang2011automatic}
M.~Wang and X.~Wang, ``Automatic adaptation of a generic pedestrian detector to
  a specific traffic scene,'' in \emph{IEEE Conference on Computer Vision and
  Pattern Recognition (CVPR)}, 2011, pp. 3401--3408.

\bibitem{wu2005detection}
B.~Wu and R.~Nevatia, ``Detection of multiple, partially occluded humans in a
  single image by bayesian combination of edgelet part detectors,'' in
  \emph{IEEE International Conference on Computer Vision}, 2005, pp. 90--97.

\bibitem{wu2019unsupervised}
Y.~Wu, Z.~Zhang, and G.~Wang, ``Unsupervised deep feature transfer for low
  resolution image classification,'' in \emph{Proceedings of the IEEE
  International Conference on Computer Vision Workshops}, 2019, pp. 0--0.

\bibitem{xu2019learn}
C.~Xu, K.~Qiu, J.~Fu, S.~Bai, Y.~Xu, and X.~Bai, ``Learn to scale: Generating
  multipolar normalized density map for crowd counting,'' \emph{arXiv preprint
  arXiv:1907.12428}, 2019.

\bibitem{xu2019adversarially}
W.~Xu, S.~Keshmiri, and G.~Wang, ``Adversarially approximated autoencoder for
  image generation and manipulation,'' \emph{IEEE Transactions on Multimedia},
  vol.~21, no.~9, pp. 2387--2396, 2019.

\bibitem{yao2017deep}
H.~Yao, K.~Han, W.~Wan, and L.~Hou, ``Deep spatial regression model for image
  crowd counting,'' \emph{arXiv preprint arXiv:1710.09757}, 2017.

\bibitem{zhang2018crowd}
L.~Zhang, M.~Shi, and Q.~Chen, ``Crowd counting via scale-adaptive
  convolutional neural network,'' in \emph{2018 IEEE Winter Conference on
  Applications of Computer Vision (WACV)}.\hskip 1em plus 0.5em minus
  0.4em\relax IEEE, 2018, pp. 1113--1121.

\bibitem{zhang2016single}
Y.~Zhang, D.~Zhou, S.~Chen, S.~Gao, and Y.~Ma, ``Single-image crowd counting
  via multi-column convolutional neural network,'' in \emph{Proceedings of the
  IEEE conference on computer vision and pattern recognition}, 2016, pp.
  589--597.

\end{thebibliography}

\balance

\end{document}